\documentclass[letterpaper]{article}
\usepackage{aaai2026}
\usepackage{times}
\usepackage{helvet}
\usepackage{courier}
\usepackage[hyphens]{url}
\usepackage{graphicx}
\usepackage{comment}
\usepackage{natbib}
\usepackage{caption}
\usepackage{amsmath,amssymb,amsfonts}
\usepackage{booktabs}
\usepackage{multirow}
\usepackage{textcomp}

\title{ARGUS: Attention-Guided Transformers for Scalable Person Identification Using Wi-Fi Telemetry}
\author{
Nayan Sanjay Bhatia\textsuperscript{1},
Pranay Kocheta\textsuperscript{2},
Yuhan Li\textsuperscript{2},
Katia Obraczka\textsuperscript{1}
}
\affiliations{
\textsuperscript{1}Department of Computer Science and Engineering, University of California, Santa Cruz, CA, USA\\
\textsuperscript{2}Independent Researcher, USA\\
nbhatia3@ucsc.edu, pranayko021@gmail.com, yuhanli27t@gmail.com, katia@soe.ucsc.edu
}

\begin{document}

\maketitle

\begin{abstract}
Passive, device-free person identification offers an alternative to camera- and wearable-based biometrics, yet existing wireless approaches rely largely on gait or activity cues and are rarely evaluated at scale. In this paper, we present \emph{Argus}, a passive Wi-Fi sensing system that identifies people from commodity Channel State Information (CSI) without requiring an attached device or a prescribed motion. Argus converts short CSI spans into compact \emph{statgrams}: statistical maps built from the channel views available on a given device. A lightweight decoder-only Transformer then reads coarse statgram patches as tokens, and segment-level logit aggregation combines evidence over time. On a 154-subject CSI dataset evaluated with a strict physical-segment split, Argus reaches $78.88\% \pm 1.62\%$ Top-1 accuracy on 6-second windows and $84.85\% \pm 1.31\%$ after aggregating 19 overlapping windows over a 60-second segment; Top-3 and Top-5 reach $98.61\%$ and $99.26\%$. For a 60-second statgram, Argus improves over a raw-CSI Transformer baseline by 7.75 points while using $4.4\times$ fewer FLOPs per window. Attention-guided compression preserves full single-window accuracy with only half of the EHealth patches. On WiMANS, a multi-user benchmark across three rooms and two Wi-Fi bands, Argus remains within 1.23 percentage points of the strongest per-configuration baselines on average while using $27\times$ fewer inference FLOPs. These results show that compact CSI statistics can scale passive identification while also exposing deployment limits in open-set rejection and cross-room transfer.
\end{abstract}

\section{Introduction}
Accurate and efficient person identification is essential for access control, personalized services, and mission-critical applications. Passwords and passkeys can be shared or stolen, while camera-based facial and gait recognition introduces privacy concerns and remains sensitive to spoofing, ambient conditions, occlusion, viewpoint, and sensor placement~\cite{Schroff2015FaceNet, Deng2019Arcface, Tolosana2020Deepfakes, Sepas-Moghaddam2021DeepGaitSurvey}. These limitations motivate low-cost, contact-less identification methods that can operate without cameras or wearable devices. 

Using telemetry from commodity Wi-Fi provides a promising low-cost alternative due to Wi-Fi's ubiquity and availability in homes, offices, and public spaces. In particular, Wi-Fi's Channel State Information (CSI) describes how wireless signals change as they propagate through an environment, capturing attenuation, reflections, and phase variations influenced by the human body. %However, CSI-based identification remains challenging. Raw CSI streams are noisy and high-dimensional, vary with room geometry, hardware, and device placement, and often require gait or other deliberate movements. Moreover, closed-set accuracy alone provides an incomplete view of system performance. A practical identification system should reveal which signal components contribute to its decisions, support computationally efficient inference, and expose failures caused by unfamiliar users or changes in the sensing environment. 

However, CSI-based identification remains challenging. Raw CSI streams are noisy and high-dimensional, and carry synchronization-induced phase offsets \cite{Diaz2023PhaseProcessing} that must be calibrated before use. They also vary with room geometry, hardware, and device placement, a domain-shift problem that cross-domain systems such as CrossSense \cite{Zhang2018CrossSense} address explicitly. Most existing systems further depend on gait or other deliberate movement, as in WiWho \cite{Zeng2016WiWho} and NeuralWave \cite{Pokkunuru2018NeuralWave}, and Transformer methods that target resting subjects have so far been evaluated mainly on small cohorts \cite{Avola2025Transformer,Wei2025WiFiSurvey}. Moreover, closed-set accuracy alone provides an incomplete view of system performance. A practical identification system should reveal which signal components contribute to its decisions, support computationally efficient inference, and expose failures caused by unfamiliar users \cite{Scheirer2013Toward} or changes in the sensing environment. The Related Work section expands on each of these limitations and on the methods proposed to address them.

We introduce \emph{Argus}, a passive person identification system based on Wi-Fi CSI telemetry. 
%built around a compact statistical representation. 
Argus takes a short window of raw CSI data and turns it into a compact summary map, called a statgram. Statgrams keep separate features of the wireless signal, such as amplitude and phase, and summarizes how each view changes over time and across subcarriers using statistics like mean, standard deviation, percentiles, and energy (Appendix Table~\ref{tab:stat-rows}). A decoder-only transformer processes coarse statgram patches as tokens and predicts identity from the final sequence representation.

% For longer recordings, Argus averages the logits of overlapping windows to produce a more stable segment-level prediction. 

We evaluated Argus using two different CSI datasets: (1) the 154-participant EHealth CSI dataset~\cite{10177905} and (2) the WiMANS dataset~\cite{Huang2024WiMANS}, a public benchmark containing multi-user Wi-Fi sensing data collected across multiple environments. Our evaluation addresses three questions. First, can statgrams support large-cohort identification under a subject-stratified split that prevents leakage between neighboring windows? Second, which regions of the statgram contain identity-relevant information, and how much of the representation can be removed without substantially reducing accuracy? Third, does the statgram-based design extend beyond single-person closed-set identification to multi-user recognition under changes in room configuration and Wi-Fi frequency band? %We evaluate Argus on the 154-participant EHealth CSI dataset~\cite{Huang2024WiMANS} and on WiMANS~\cite{10177905}, a public benchmark containing multi-user Wi-Fi sensing data collected across multiple environments.
Our contributions are as follows:
\begin{itemize}
\item We present the \emph{Argus} design and architecture in detail 
%, a passive Wi-Fi CSI system for scalable person identification. 
%\item We 
and conducted a thorough evaluation of its pipeline. We showed that, in the 154-subject EHealth dataset, Argus improves from $78.88\% \pm 1.62\%$ Top-1 accuracy on 6-second windows to $84.85\% \pm 1.31\%$ when compared to 60-second segments, with Top-5 reaching $99.26\%$.
\item We show that compact representation is more important than simply extending raw temporal context. At matched 60-second evidence, Argus outperforms a raw-CSI Transformer baseline by 7.75 points while using $4.4\times$ fewer FLOPs per window.
\item We introduce validation-only attention-occlusion ranking for CSI statgrams. Retaining the top 4 of 8 EHealth patches preserves full single-window accuracy, and retaining 10 of 15 WiMANS patches costs only 0.7 points on average.
\item We evaluate deployment limits rather than treating closed-set accuracy as sufficient. Adjacent-ID analysis reveals structured residual errors, open-set rejection remains moderate, and WiMANS cross-room experiments show that zero-shot transfer is weak even when few-shot mixed training helps.
\end{itemize}

\section{Related Work}
\noindent\textbf{CSI-based person identification.}
Traditional biometric systems, such as vision-based recognition and radar, attain high accuracy but struggle with line-of-sight requirements, privacy concerns, or the need for specialized hardware~\cite{Schroff2015FaceNet, Wang2024FRTPrivacy, Vandersmissen2018IndoorRadar, Buyukakkaslar2024RadarSurvey}. Radar-based identification avoids lighting dependence by analyzing RF reflections, and most frameworks exploit micro-Doppler signatures produced by limb motion during walking~\cite{Vandersmissen2018IndoorRadar}. High-resolution millimeter-wave radar can even resolve respiration and heartbeat-scale displacements, enabling contactless biometrics for stationary subjects, but these systems still require dedicated sensing hardware and radar-specific deployment geometries~\cite{Buyukakkaslar2024RadarSurvey}. Commodity Wi-Fi CSI provides a passive alternative by capturing multipath amplitude and phase variations caused by human presence~\cite{Ma2019CSISurvey}. Initial CSI identification systems used feature-engineered classifiers, and later systems adopted CNNs, LSTMs, and other deep models for identity classification~\cite{Ding2020Wihi, Pokkunuru2018NeuralWave, Wei2025WiFiSurvey}. These systems show that CSI carries biometric information, but many rely on prominent gait- or activity-induced channel variation
\cite{Zeng2016WiWho,Pokkunuru2018NeuralWave}. This limits their applicability to stationary identification, where the available signal is much weaker.

Scaling Wi-Fi identification across rooms and people is also difficult. Recent work has explored cross-domain sensing and few-shot identity-similarity learning to improve domain adaptability~\cite{Zhang2018CrossSense, WangSimID2025}. These methods are important because Wi-Fi channels are strongly tied to the room. However, relying on large body motion remains a limitation. Specialized RF studies show that physiological micro-motions can perturb wireless channels even when a person is at rest~\cite{Adib2015VitalRadio, Kocheta2025PulseFi}. Argus is motivated by this observation: the signal may be weak, but it can become usable if the representation is compact, stable, and evaluated at scale.

\noindent\textbf{Transformers and interpretability.}
CSI processing has moved toward Transformer architectures because they can model long-range temporal dependencies and combine amplitude and phase perturbations~\cite{Li2021CSITransformer, Quy2025CSIActivity}. Recent dual-branch Transformers process CSI amplitude and phase for person identification from resting individuals~\cite{Avola2025Transformer}, but are mostly limited to small cohorts. Current identification research continues to identify scale, environment shift, and open-set recognition as open problems~\cite{Chen2023WiFiSurvey, Wei2025WiFiSurvey}. In parallel, pruning and architecture-optimization methods show that large attention models can often be reduced without losing much accuracy~\cite{Michel2019Sixteen, Kwon2022FastPruning, Youm2025CSIPruning}. Attention mechanisms also provide a way to inspect which signal regions affect decisions~\cite{Abnar2020Rollout, Yang2022AGait}. Argus combines these ideas by using attention-guided occlusion to rank statgram patches and then testing whether those patches can be retained as a compressed input.

\section{Argus}\label{sec:argus}
Argus is a generalized pipeline for passive identification from CSI streams. The method does not assume a particular number of radios, antennas, channel bins, or enrolled users. Instead, it assumes only that a recording can be represented as a time-ordered stream of CSI measurements and that each measurement can be transformed into one or more channel views, such as amplitude, calibrated phase, or another stable function of the complex CSI.
\begin{figure}[!t]
\centering
\includegraphics[width=\columnwidth]{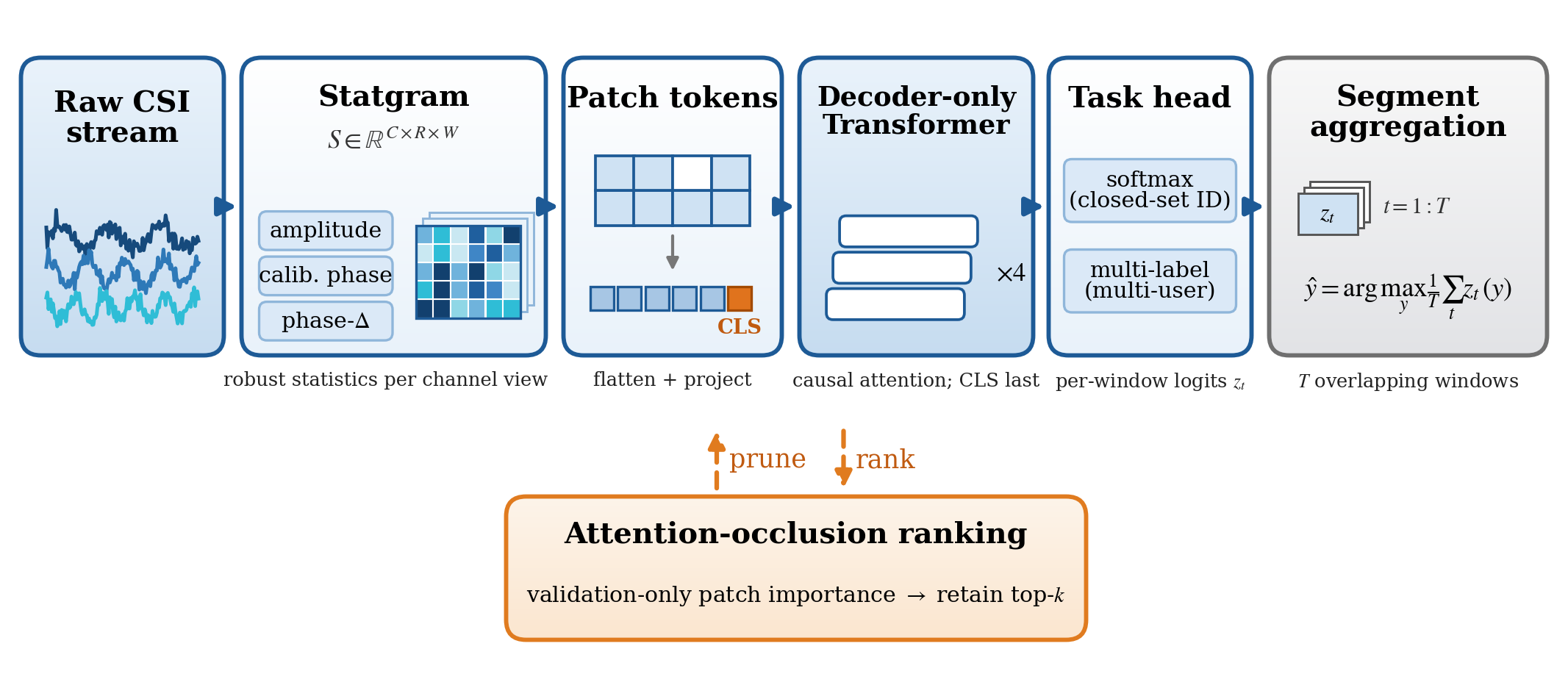}
\caption{Argus system architecture.}
\label{fig:argus-arch}
\end{figure}

\subsection{Statgram Representation}
Raw CSI packet streams contain outliers, packet-level jitter, hardware-dependent scaling, and redundant high-frequency variation. Directly modeling every packet therefore gives the model many tokens but does not guarantee a stable biometric representation. Argus converts short CSI spans into \emph{statgrams}. Let a window contain complex CSI
\[
H \in \mathbb{C}^{T \times K},
\]
with $T$ packets and $K$ subcarriers. For a real-valued matrix $X$, define robust normalization
\[
\rho(X)=\frac{X-\operatorname{median}(X)}{\max(\operatorname{IQR}(X),\epsilon)},
\]
where the median and interquartile range are computed over the whole window and $\epsilon=10^{-4}$. Let $g_W(\cdot)$ average adjacent subcarriers into $W$ grouped bins. In the EHealth implementation, the amplitude view is
\[
A = g_W\!\left(\rho(\log(1+|H|))\right).
\]
For phase, Argus unwraps along the subcarrier axis and removes a packet-wise endpoint line,
\[
\hat{\Phi}_{t,k}=\operatorname{unwrap}_k(\angle H_{t,k}),\quad
a_t=\frac{\hat{\Phi}_{t,K}-\hat{\Phi}_{t,1}}{K-1},
\]
\[
\Phi_{t,k}=\hat{\Phi}_{t,k}-\bigl(a_t(k-1)+\hat{\Phi}_{t,1}\bigr),\quad
P=g_W(\rho(\Phi)).
\]
This linear phase detrending follows the common CSI observation that the measured phase contains synchronization-related offsets that are approximately linear in the subcarrier index~\cite{Diaz2023PhaseProcessing}. The phase-delta view is the first temporal difference of the grouped phase,
\[
D_{t,w}=P_{t,w}-P_{t-1,w},\quad D_{1,w}=0.
\]
For each view $X^{(v)} \in \{A,P,D\}$, Argus applies statistic rows $s_r$ over time for each grouped bin, so
\[
S_{v,r,w}=s_r\!\left(X^{(v)}_{1:T,w}\right).
\]
Frequency rows use temporal real-FFT power after subtracting the temporal mean, discarding the DC bin, and splitting the remaining bins into low, middle, and high thirds. Appendix Table~\ref{tab:stat-rows} enumerates the statistic rows. These statistics are stacked into a tensor of the form
\[
S \in \mathbb{R}^{C \times R \times W},
\]
where $C$ is the number of available views, $R$ is the number of statistic rows, and $W$ is the width induced by the capture geometry after grouping or alignment. The important point is that $C$, $R$, and $W$ are implementation choices, not assumptions of the method. A different radio or task can change the statgram shape while keeping the same learning pipeline.

\subsection{Patch Transformer}
The statgram is divided into coarse rectangular patches. Each patch is flattened and linearly projected into a token. A learnable CLS token is placed at the end of the sequence and processed by a lightweight decoder-only Transformer. Because the CLS token is last, causal attention still allows it to attend to every statgram patch. The final CLS embedding is passed to an output head chosen for the task: a softmax head for closed-set single-person identification, or independent binary logits for multi-user identification.

This design is intentionally small. The goal is not to build a large sequence model over raw packets, but to make the input stable enough that a short token sequence is sufficient. Segment-level inference then averages logits across overlapping windows from the same physical recording:
\[
\hat{y} = \arg\max_y \frac{1}{T}\sum_{t=1}^{T} z_t(y),
\]
where $z_t(y)$ is the logit for identity $y$ in window $t$. This aggregation is deterministic and improves robustness without changing the trained model.

\subsection{Attention-Occlusion Compression}
Argus also uses the trained model to identify which statgram patches matter. For each validation window, we first compute the unmasked prediction. We then replace one patch at a time with its training-set mean value and measure the drop in true-class probability. Patch importance is the average validation drop, with accuracy drop used as a secondary signal. Test-time compression is evaluated only after this validation ranking is fixed. We compare ranked mean-fill, ranked zero-fill, and random mean-fill controls so that any compression gain cannot be explained merely by keeping some arbitrary subset of patches.

\section{Experimental Methodology}\label{sec:experiments}
\subsection{Datasets}
We evaluate Argus on two CSI datasets. 

% The EHealth CSI dataset~\cite{10177905} contains CSI from 154 participants collected in a controlled 3 m $\times$ 4 m indoor room. Participant ages range from 18 to 64 years (mean 22.38), heights from 152 to 198 cm, and weights from 40 to 116 kg, producing substantial overlap in physical characteristics such as body mass index. Each participant completed 17 one-minute positions or activities. Positions 1, 2, and 4 to 13 as stationary or breathing-variation recordings, and positions 3 and 14--17 as posture-transition, locomotion, or activity recordings. The released table contains 2,618 physical recordings stored as 500 CSI rows per 60-second recording (a roughly 136 ms ping interval, i.e., $\approx$7.4 Hz effective sampling). The capture used a 5 GHz router, a laptop client, and a single-antenna Raspberry Pi 4B probe running NEXMON firmware~\cite{nexmon:project}. For this dataset, each 50-row window, approximately 6 seconds, is represented as a $3 \times 20 \times 64$ statgram from amplitude, calibrated phase, and phase-delta statistics; adjacent windows use a 25-row stride. The model uses a $10 \times 16$ patch size, yielding a $2 \times 4$ grid of 8 patch tokens plus one CLS token.
The EHealth dataset \cite{10177905} contains CSI from 154 participants collected in a controlled $3\,\text{m} \times 4\,\text{m}$ room, providing a large-cohort benchmark for Wi-Fi-based identification. Participants varied in age, height, and weight, with substantial overlap in physical characteristics. Each completed 17 standardized 60-second positions or activities spanning static poses, breathing variations, and limited in-place motion. Because 14 activities were static, the dataset is well suited for passive identification of stationary subjects. CSI was collected over 256 subcarriers, 234 usable, using a 5 GHz router on channel 36 with 80 MHz bandwidth, a laptop client, and a single-antenna Raspberry Pi 4B running NEXMON. The router and laptop were placed on opposite sides of the participant, while the probe was equidistant from both, with all devices approximately 1 m away. Each 50-row window, or roughly 6 seconds, is converted into a $3 \times 20 \times 64$ statgram using amplitude, calibrated phase, and phase-delta statistics, with a 25-row stride. A $10 \times 16$ patch size produces a $2 \times 4$ grid of eight patch tokens plus one CLS token.

WiMANS contains 11,286 three-second recordings across classroom, meeting-room, and empty-room environments, at 2.4 GHz and 5 GHz~\cite{Huang2024WiMANS}. Each recording contains zero to five of six enrolled users. The released preprocessing stores CSI magnitude, so we re-extract the complex CSI from the raw traces and construct per-link amplitude and calibrated phase views. The final WiMANS statgram has shape $18 \times 20 \times 30$ and uses $4 \times 10$ patches, yielding 15 patch tokens plus the CLS token. The output head is replaced with six independent logits trained with weighted binary cross-entropy.

\subsection{Splits, Baselines, and Metrics}
For EHealth, we use a subject-stratified physical-segment split. Each participant contributes 11 training, 3 validation, and 3 test segments, yielding 462 physical test segments and preventing neighboring windows from the same recording from crossing splits. No demographic or physiological metadata is used as input. For WiMANS, we follow the benchmark's 80/20 split and exact-match multi-label metric, under which a recording counts as correct only if all six per-user predictions are simultaneously correct.

On EHealth, we compare Argus with THAT, a two-stream raw-CSI Transformer adapted to 154-way identification~\cite{Li2021CSITransformer}; a CNN-BiLSTM with convolutional feature extraction, bidirectional recurrence, and a softmax identity head; and a full Argus encoder-decoder using bidirectional self-attention and a single cross-attention query token. The encoder--decoder uses the same statgram input, patching, width, training budget, and five seeds, but has 4.2M parameters versus 1.9M for the decoder-only model. On WiMANS, we compare with the published LSTM, CNN-1D, CNN-2D, CLSTM, ABLSTM, and THAT baselines. Because Argus re-extracts complex CSI to include calibrated phase, these are published-reference rather than fully matched-input comparisons. We report EHealth Top-$k$ accuracy, WiMANS exact-match accuracy, attention-occlusion compression, open-set rejection, calibration, and FLOPs. For aggregated results, x$N$ averages logits over the first $N$ overlapping 6-second windows from each 60-second segment using a 3-second stride; x19 uses all 19 windows. The main Argus model uses $d_{\text{model}}=192$, four decoder blocks, six heads, 0.15 dropout, 1.90M parameters, and 33.6 MFLOPs per EHealth window.

For open-set evaluation, a deterministic 20\% subset of participant IDs (31 of 154) is withheld from training and treated as unknown. The model is retrained from scratch on the remaining 123 enrolled IDs. Known-user accuracy is measured only on enrolled-user test recordings, while unknown rejection is evaluated with maximum softmax confidence and normalized negative entropy. We also compute 15-bin expected calibration error (ECE) on known-user predictions and sweep confidence thresholds to expose the trade-off among known-user acceptance, unknown false acceptance, and accuracy among accepted known users.

\section{Results}\label{sec:results}
\subsection{Large-Cohort EHealth Identification}
\begin{figure}[t]
\centering
\includegraphics[width=0.75\columnwidth]{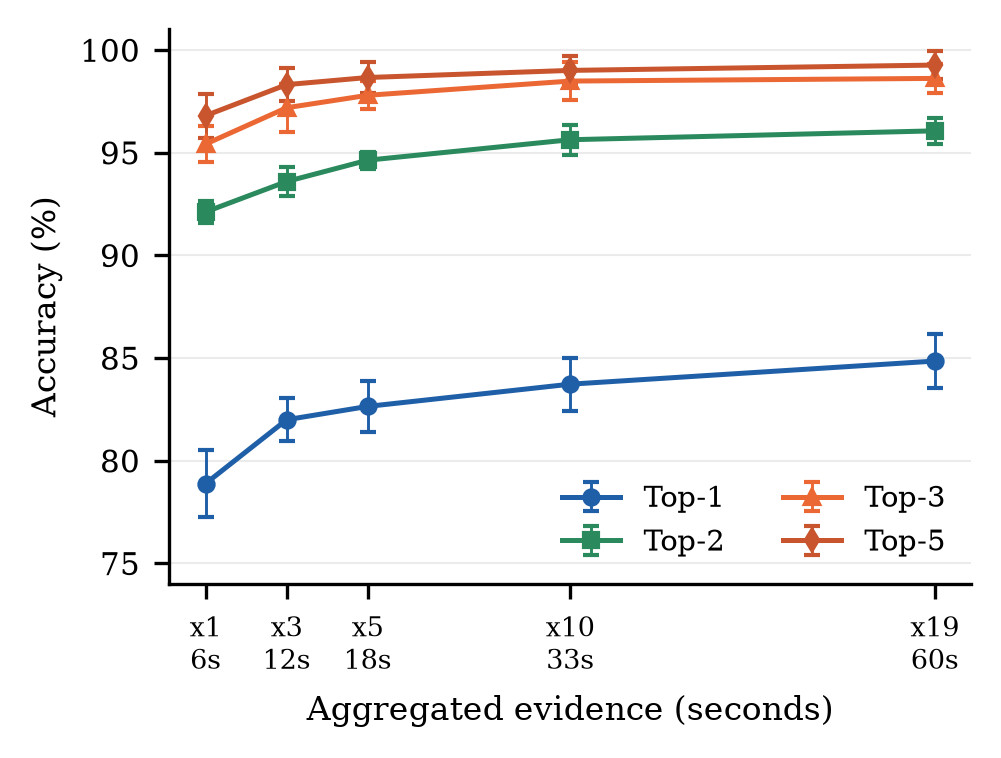}
\caption{EHealth Top-$k$ accuracy versus aggregated evidence.}
\label{fig:eh-agg}
\end{figure}

Table~\ref{tab:ehealth-main} shows the main EHealth result. A single 6-second statgram reaches $78.88\% \pm 1.62\%$ Top-1 accuracy. Aggregating logits over 19 overlapping windows from a 60-second physical segment raises Top-1 to $84.85\% \pm 1.31\%$, while Top-3 and Top-5 reach $98.61\%$ and $99.26\%$ (Figure~\ref{fig:eh-agg}). This is important because the application does not always need an exact single-shot match; a short candidate list with very high recall is useful for a secondary verification step.

\begin{table}[t]
\centering
\small
\caption{EHealth closed-set identification accuracy.}
\label{tab:ehealth-main}
\begin{tabular}{lrrrr}
\toprule
Model & Sec. & Top-1 & Top-3 & Top-5 \\
\midrule
Statgram x1 & 6 & $78.88{\pm}1.62$ & 95.42 & 96.80 \\
Statgram x5 & 18 & $82.64{\pm}1.23$ & 97.79 & 98.66 \\
Statgram x19 & 60 & $84.85{\pm}1.31$ & 98.61 & 99.26 \\
Statgram-EncDec x1 & 6 & $79.15{\pm}1.37$ & 94.29 & 95.89 \\
Statgram-EncDec x19 & 60 & $85.41{\pm}1.78$ & 98.18 & 98.79 \\
THAT raw x1 & 6 & $82.37{\pm}0.57$ & 96.76 & 98.25 \\
THAT raw x5 & 18 & $78.04{\pm}0.92$ & 96.73 & 98.59 \\
THAT raw x19 & 60 & $77.10{\pm}0.47$ & 97.45 & 99.13 \\
CNN-BiLSTM & 6 & $75.08{\pm}0.61$ & -- & -- \\
CNN-BiLSTM & 60 & $65.57{\pm}2.09$ & -- & -- \\
\bottomrule
\end{tabular}
\end{table}

The raw-CSI THAT model performs best on a single 6-second window, exceeding Argus by 3.49 points, but its Top-1 accuracy falls as the context grows. In contrast, Argus improves with accumulated evidence, reaching 84.85\% at 60 seconds and outperforming THAT by 7.75 points. Argus is also more efficient, requiring 33.6 MFLOPs per window compared with 149 MFLOPs for THAT. A statgram forward pass takes 1.21 ms median on Apple Silicon, resulting in approximately 23 ms for an x19 segment-level decision.
The two additional comparisons reinforce this pattern. CNN-BiLSTM reaches $75.08\% \pm 0.61\%$ Top-1 accuracy on 6-second raw-CSI windows but falls to $65.57\% \pm 2.09\%$ over 60 seconds, mirroring THAT’s degradation with longer raw context. In contrast, the encoder-decoder statgram model closely matches the decoder-only variant, reaching $79.15\% \pm 1.37\%$ at x1 and $85.41\% \pm 1.78\%$ at x19, a difference within seed variance, despite using more than twice as many parameters. These results suggest that the statgram representation is the main source of the aggregation gains, rather than the choice between decoder-only, encoder-decoder, or recurrent architectures.

\begin{figure}[t]
\centering
\includegraphics[width=0.75\columnwidth]{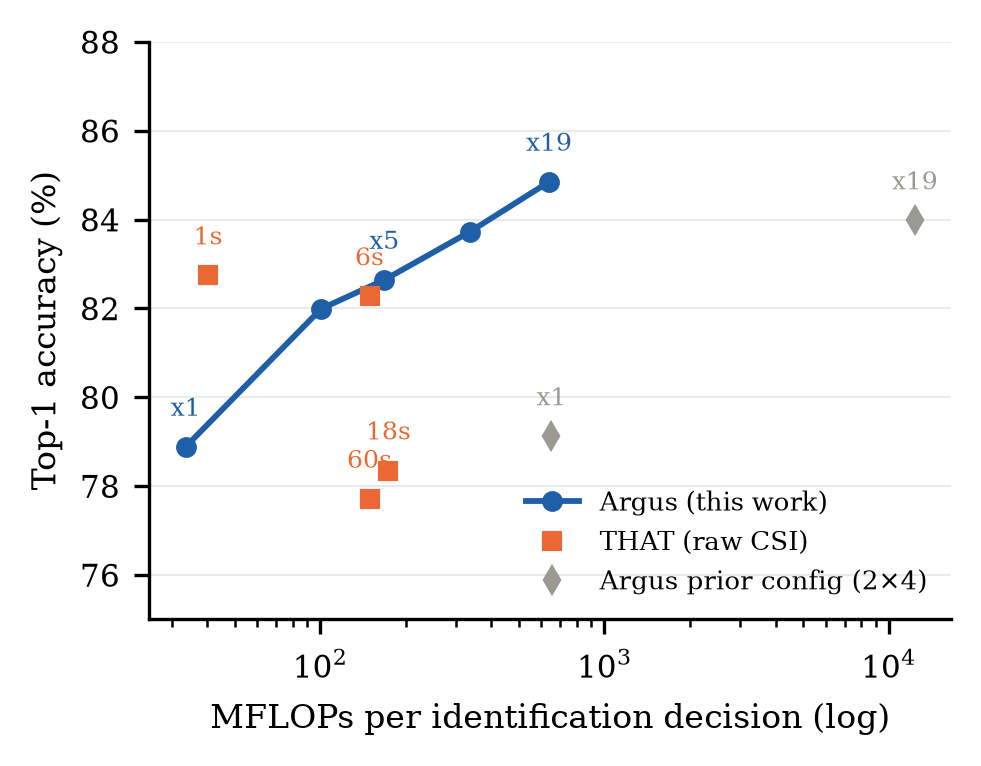}
\caption{Accuracy-compute frontier on EHealth.}
\label{fig:eh-frontier}
\end{figure}

\subsection{Channel Ablation and Compression}
\begin{figure}
\centering
\includegraphics[width=0.75\columnwidth]{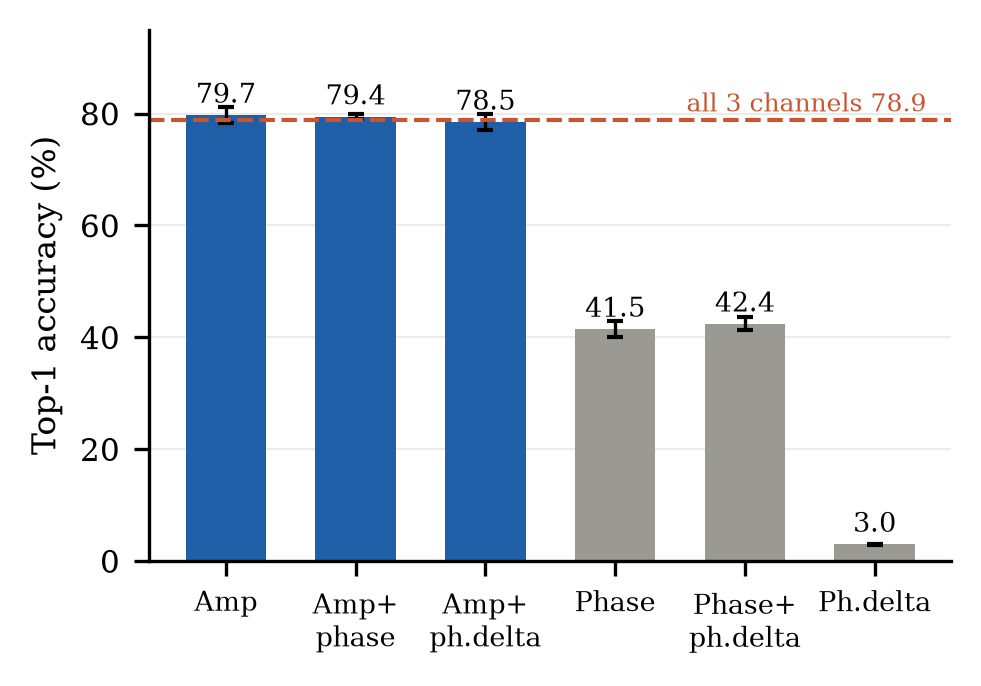}
\caption{Single-window Top-1 accuracy under amplitude, phase, and phase-delta channel ablations.}
\label{fig:eh-chan}
\end{figure}

The EHealth channel ablation shows that amplitude carries most of the discriminative signal in the single-radio setting (Figure~\ref{fig:eh-chan}). Amplitude-only reaches $79.72\% \pm 1.49\%$ single-window Top-1, statistically indistinguishable from the full three-channel model. Phase-only drops to $41.47\%$, and phase-delta-only drops to $3.00\%$. This does not mean that phase is useless in general. Rather, it shows that phase calibration and channel selection need to be treated as first-class design choices. On WiMANS, calibrated per-link phase helps most on the noisier 2.4 GHz configurations.

\begin{figure}[t]
\centering
\includegraphics[width=0.75\columnwidth]{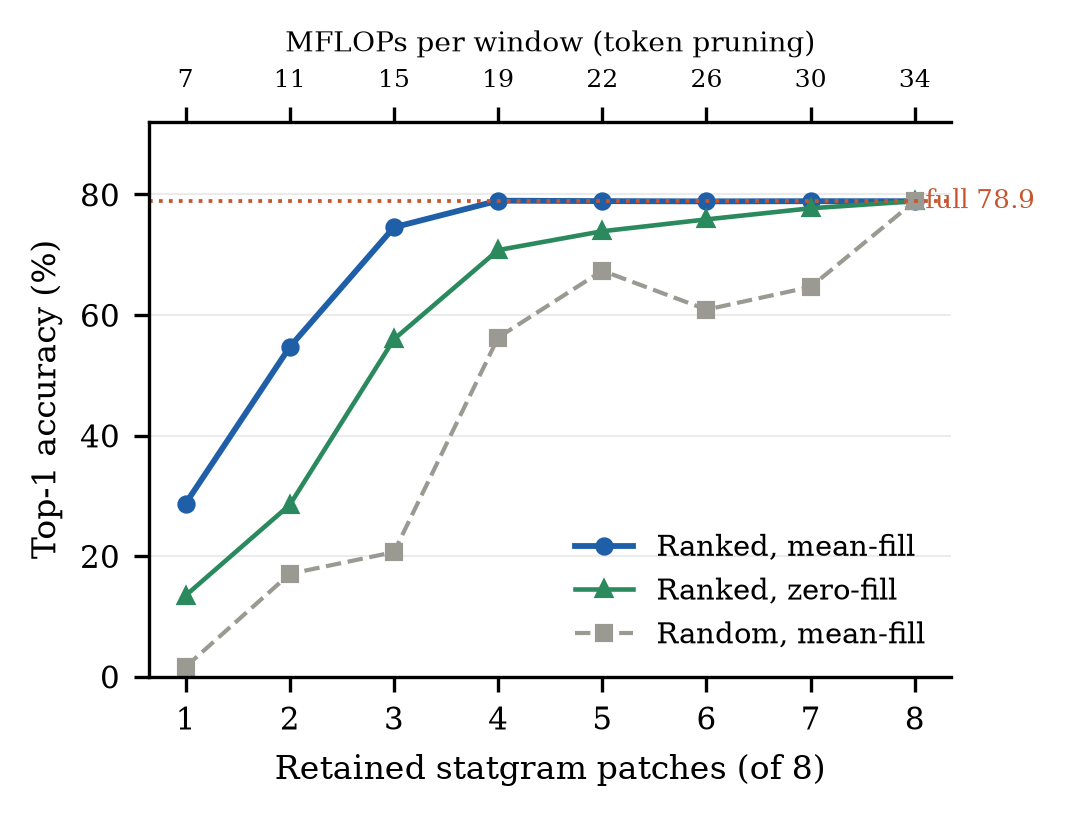}
\caption{Input compression on the 8-patch EHealth statgram.}
\label{fig:eh-ret}
\end{figure}

\begin{figure}[t]
\centering
\includegraphics[width=0.75\columnwidth]{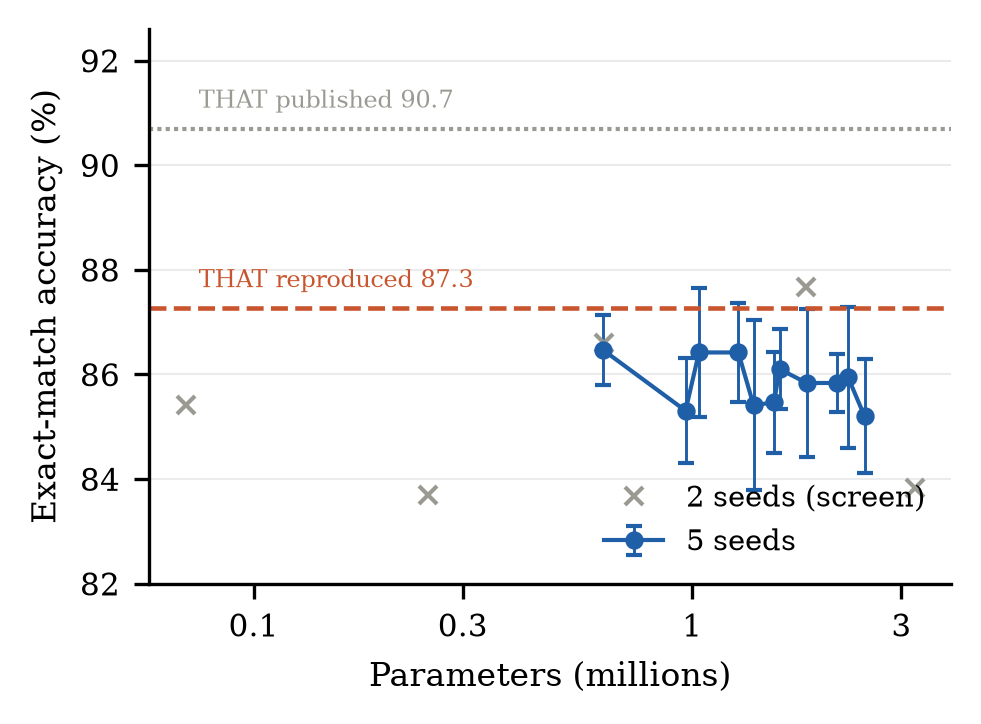}
\caption{Accuracy versus model size on classroom 2.4 GHz.}
\label{fig:capacity}
\end{figure}

Attention-guided occlusion shows that identity evidence is spatially concentrated. Retaining the top 4 of 8 validation-ranked patches with training-mean fill preserves full single-window accuracy ($78.97\%$ versus $78.88\%$), corresponding to 18.6 MFLOPs per window under token pruning and accuracy declines only below 3 retrained patches (Figure~\ref{fig:eh-ret}). Random retention is much worse at the same budgets, and zero-fill recovers more slowly than mean-fill which indicates that the ranking selects meaningful patches rather than simply reducing input size. The model retrained on only the top-ranked half of the statistic rows (a $10 \times 64$ input) reaches $79.35\% \pm 1.29\%$, slightly above the full-input reference, while column-only reductions cost more (top-half subcarrier columns: $76.12\% \pm 1.18\%$), suggesting the statistic rows are more redundant than the subcarrier groups. On WiMANS, retaining 10 of 15 ranked patches costs only 0.7 points on average and accuracy collapses below 5 patches. When testing model capacity on WiMANS, accuracy is statistically flat across a $4\times$ parameter range (0.6M--2.5M) and an 11M-parameter variant fails to generalize entirely, so the compact input, not model size, is the binding design constraint, seen in Figure~\ref{fig:capacity}.

\subsection{Failure Modes and Open-Set Behavior}
The remaining EHealth errors are not random. At single-window inference, $61.9\% \pm 0.6\%$ of Top-1 errors are adjacent participant IDs. A random output-permutation null gives only 1.31\% adjacent errors on average, while random-label retraining preserves high adjacency only when predictions are mapped back to the original participant order. This suggests that residual errors reflect collection order, session structure, participant similarity, or another latent dataset factor. The result is useful, but it also means that future evaluation should randomize collection order or test across additional sessions.

\begin{figure}[t]
\centering
\includegraphics[width=0.75\columnwidth]{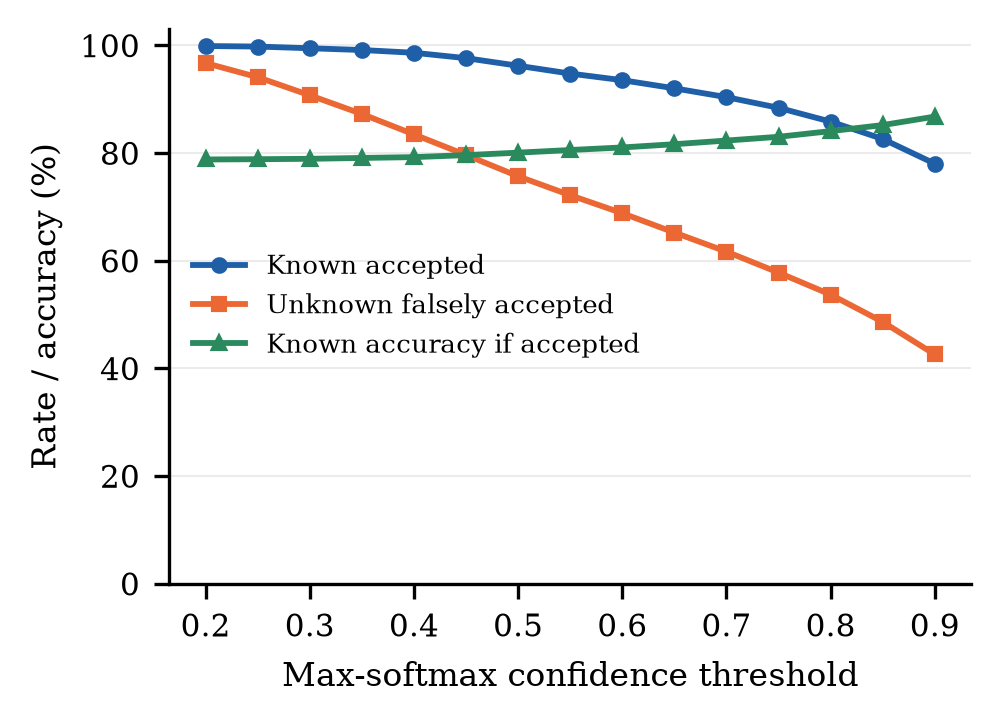}
\caption{Open-set threshold trade-off with 31 withheld identities.}
\label{fig:eh-open}
\end{figure}

Open-set rejection is also not solved. With 31 of 154 participants withheld from training, Argus reaches 78.69\% known-user accuracy. Unknown-user rejection gives AUROC 0.744 using maximum confidence and 0.754 using negative entropy. The threshold sweep in Figure~\ref{fig:eh-open} further shows that reducing unknown false accepts comes at the cost of rejecting more known users. These values are useful diagnostics, but they are not strong enough for a stand-alone biometric gate. Argus should therefore be viewed as a closed-set identifier or high-recall shortlist generator unless it is combined with a stronger verification objective.

\subsection{WiMANS Multi-User Evaluation}
WiMANS tests whether the same representation family remains useful when the task changes from single-person closed-set identification to multi-user exact-match recognition. Across the nine environment-band configurations, Argus reaches 95.07\% mean exact-match accuracy, within 1.23 points of the strongest per-configuration baselines on average and within one point on five configurations (Table~\ref{tab:wimans_results}). The remaining gaps are concentrated in the empty-room 5 GHz and empty-room dual-band settings. We retrain all baselines for a like-for-like check on the same hardware and pipeline. When retraining on the same hardware, the difference between Argus and the best baseline drops to 0.97 points on average which is within seed variance. Full retrained baseline results can be found in appendix.

Two post-hoc observations are worth reporting (excluded from Table~\ref{tab:wimans_results}). First, on the cleaner 5 GHz band the phase channels are slightly counter-productive: an amplitude-only variant of the final model reaches $99.58\% \pm 0.27\%$ on meeting-room 5 GHz ($-0.41$ versus THAT's 99.99\%), suggesting band-adaptive channel selection as a simple deployment heuristic. Second, a tied-optimal $5 \times 10$ patch variant reaches $95.65\% \pm 0.88\%$ on meeting-room 2.4 GHz. We also report a negative phase result: encoding cross-antenna conjugate-product phase differences as cosine and sine channels reduced accuracy by 3--4 points in every combination we tested, and phase-only models trailed amplitude-only models by roughly 9 points.

\begin{table*}[!t]
\centering
\footnotesize
\setlength{\tabcolsep}{2.6pt}
\begin{tabular}{llcccccccr}
\toprule
Environment & Band & LSTM & CNN-1D & CNN-2D & CLSTM & ABLSTM & THAT & Argus (ours) & Gap \\
\midrule
Classroom & 2.4 & $80.74{\pm}1.43$ & $84.80{\pm}0.80$ & $89.97{\pm}0.66$ & $90.16{\pm}0.70$ & $88.81{\pm}1.36$ & $\mathbf{90.69{\pm}0.51}$ & $89.81{\pm}1.36$ & $-0.88$ \\
Meeting & 2.4 & $89.52{\pm}1.11$ & $92.89{\pm}0.26$ & $\mathbf{96.39{\pm}0.81}$ & $94.56{\pm}0.49$ & $94.80{\pm}0.55$ & $94.77{\pm}0.59$ & $94.69{\pm}0.24$ & $-1.70$ \\
Empty & 2.4 & $86.68{\pm}0.99$ & $88.14{\pm}0.46$ & $91.11{\pm}0.82$ & $92.76{\pm}0.27$ & $90.90{\pm}0.54$ & $\mathbf{93.87{\pm}0.88}$ & $92.94{\pm}0.81$ & $-0.93$ \\
\midrule
Classroom & 5 & $98.89{\pm}0.16$ & $99.47{\pm}0.21$ & $99.10{\pm}0.21$ & $\mathbf{99.71{\pm}0.19}$ & $99.55{\pm}0.29$ & $99.23{\pm}0.19$ & $98.89{\pm}0.20$ & $-0.82$ \\
Meeting & 5 & $99.73{\pm}0.17$ & $99.66{\pm}0.12$ & $99.79{\pm}0.11$ & $99.76{\pm}0.08$ & $99.76{\pm}0.08$ & $\mathbf{99.99{\pm}0.00}$ & $98.67{\pm}0.29$ & $-1.32$ \\
Empty & 5 & $95.84{\pm}0.50$ & $95.94{\pm}0.36$ & $95.31{\pm}0.36$ & $\mathbf{97.64{\pm}0.25}$ & $96.74{\pm}0.38$ & $97.08{\pm}0.29$ & $95.49{\pm}0.69$ & $-2.15$ \\
\midrule
Classroom & 2.4/5 & $86.93{\pm}0.90$ & $92.64{\pm}0.42$ & $\mathbf{94.61{\pm}0.50}$ & $93.05{\pm}0.46$ & $92.35{\pm}0.39$ & $94.29{\pm}0.51$ & $93.97{\pm}0.30$ & $-0.64$ \\
Meeting & 2.4/5 & $93.94{\pm}0.60$ & $96.64{\pm}0.21$ & $\mathbf{98.01{\pm}0.35}$ & $97.44{\pm}0.28$ & $97.84{\pm}0.17$ & $97.73{\pm}0.33$ & $97.85{\pm}0.27$ & $-0.16$ \\
Empty & 2.4/5 & $90.07{\pm}0.95$ & $93.49{\pm}0.21$ & $94.45{\pm}0.41$ & $95.29{\pm}0.38$ & $95.01{\pm}0.34$ & $\mathbf{95.80{\pm}0.40}$ & $93.31{\pm}0.68$ & $-2.49$ \\
\midrule
Mean & & 91.37 & 93.74 & 95.42 & 95.60 & 95.08 & $\mathbf{95.94}$ & 95.07 & $-1.23$ \\
\bottomrule
\end{tabular}
\caption{Exact-match identification accuracy on WiMANS for the top 6 baselines~\cite{Huang2024WiMANS} and Argus.}
\label{tab:wimans_results}
\end{table*}

\begin{comment}

\begin{figure*}[t]
\centering
\includegraphics[width=\textwidth]{wimans_main.png}
\caption{Exact-match accuracy on the nine WiMANS environment-band configurations.}
\label{fig:wimans_main}
\end{figure*}

\end{comment}

The WiMANS adaptation also exposes which design choices contributed the most. Re-extracting calibrated phase from the raw traces, conditioning statistic rows, using coarser patches, and switching to a multi-label objective all improve the hardest classroom 2.4 GHz configuration. Phase is band-dependent: it gives its largest gains at 2.4 GHz, where amplitude is noisier, and is neutral or slightly negative on cleaner 5 GHz settings. Token granularity matters as well. With roughly 1,500 training recordings per configuration, coarse 4-by-10 patches work better than fine tokenization because they reduce the effective input dimensionality.

\begin{table}[t]
\centering
\small
\caption{WiMANS accuracy/compute summary. Speedup is relative to THAT; baseline parameter and FLOP counts are the benchmark's published figures~\cite{Huang2024WiMANS}.}
\label{tab:compute-main}
\begin{tabular}{lrrrr}
\toprule
Model & Params (M) & Acc. & GFLOPs & Speedup \\
\midrule
THAT & 4.900 & $\mathbf{95.94}$ & 1.650 & $1.0\times$ \\
CLSTM & 5.391 & 95.60 & 1.791 & $0.9\times$ \\
CNN-2D & 0.893 & 95.42 & 1.691 & $1.0\times$ \\
ABLSTM & 4.268 & 95.08 & 3.208 & $0.5\times$ \\
CNN-1D & 1.916 & 93.74 & 0.516 & $3.2\times$ \\
LSTM & 1.609 & 91.37 & 0.971 & $1.7\times$ \\
MLP & 209.020 & 83.45 & 0.418 & $3.9\times$ \\
\midrule
Argus & 1.920 & 95.07 & 0.061 & $27\times$ \\
\quad{}10 patch & 1.920 & 94.37 & 0.042 & $39\times$ \\
\quad{}8 patch & 1.920 & 93.57 & 0.034 & $48\times$ \\
\bottomrule
\end{tabular}
\end{table}

The efficiency result is the strongest deployment argument. Argus uses 0.061 GFLOPs per recording, $27\times$ fewer than THAT, while giving up 0.87 points of mean accuracy to THAT and 1.23 points to the strongest per-configuration baselines. Combining patch retention with token pruning extends the speedup to $39\times$ for 10 patches and $48\times$ for 8 patches. The reason is simple: the statgram compresses thousands of packet-level measurements into a short token sequence before any attention is computed.

\subsection{Cross-Environment Transfer}
We use a leave-one-environment-out protocol within each band setting: models are trained on all recordings of the two source environments and evaluated on the held-out environment's standard 20\% test split, with best-epoch selection on a 10\% validation carve-out of the training pool. As a fraction $x \in \{1,2,3,4,5,10,15,20,25,50\}\%$ of the target environment's training split is made available (three seeds), we compare three adaptation strategies: \emph{mixed} training on the source pool plus the $x\%$ target sample, \emph{fine-tuning} a source-pretrained model on the $x\%$ target sample alone (100 epochs, learning rate $10^{-4}$), and a \emph{target-only} control trained on the $x\%$ sample with no source data.

\begin{figure*}[t]
\centering
\includegraphics[width=\textwidth]{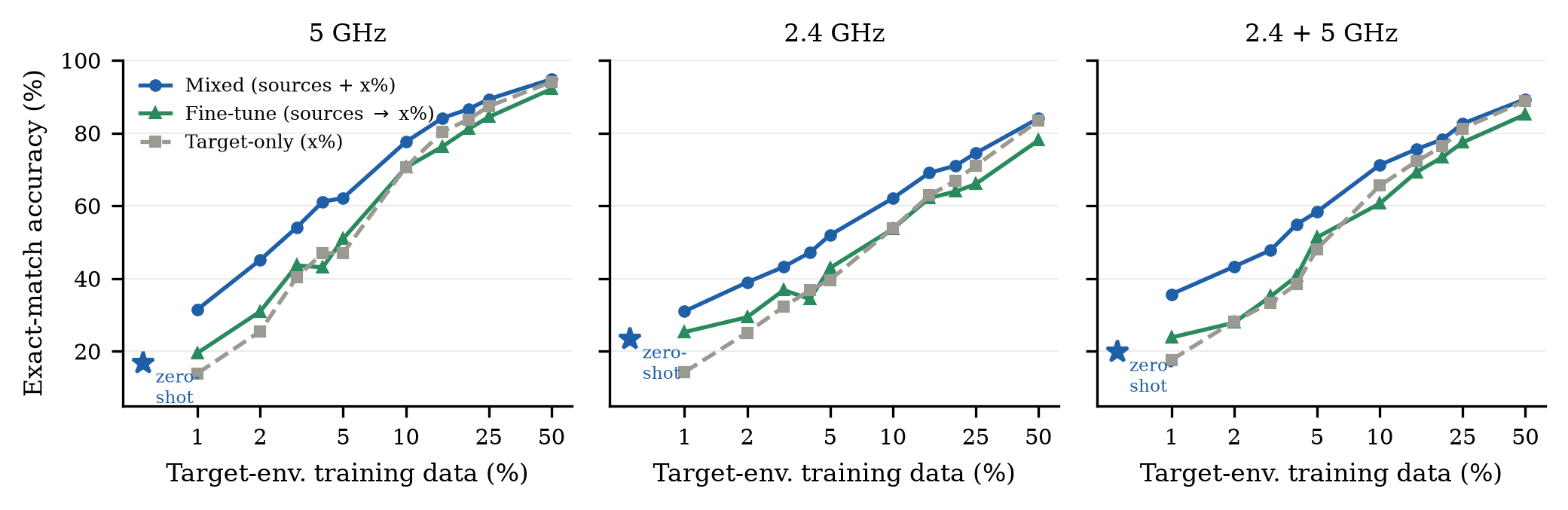}
\caption{Cross-environment adaptation on WiMANS.}
\label{fig:xenv}
\end{figure*}

Three findings stand out (Figure~\ref{fig:xenv}). First, zero-shot transfer fails: models trained on two rooms reach only 14--28\% exact match on the held-out room, showing that CSI identity signatures are dominated by room-specific multipath. Second, transfer is nonetheless real and concentrated in the few-shot regime: at 1\% target data (15 recordings), mixed training reaches 31--36\% versus 14--18\% for target-only training, a gain that persists at $+$10--15 points through 5\% target data and fades to zero by 50\%. The source environments roughly halve the target-room enrollment data needed to reach a given accuracy below the 10\% mark, and with a quarter of the target data (376 recordings) mixed training reaches 82--93\% depending on band and room. Third, and contrary to common transfer-learning practice, mixed training beats pretrain-then-fine-tune at every budget and in every band: fine-tuning on small target samples barely improves on the target-only control, indicating that the source data regularizes joint training rather than providing a directly transferable initialization. We did not tune fine-tuning hyperparameters beyond a standard recipe, so a carefully tuned variant may close part of this gap.

\section{Discussion and Future Work}
Argus' main contribution is its compact input representation using statgrams which enable identification-relevant CSI structure learnable with short token sequences. Our evaluation results show that longer raw temporal context does not necessarily yield improved performance:
%alone is not sufficient: 
the raw-CSI Transformer degrades with longer contexts, while statgram aggregation improves. This is consistent with the compression results, where a small number of ranked patches preserve most of the accuracy.

There are also clear limits. First, amplitude dominates the EHealth single-radio setting, while calibrated phase helps selectively in WiMANS. Future systems should therefore choose channel views adaptively rather than assuming that more views always help. Second, Argus has only moderate ability to detect unenrolled individuals, so it is currently better suited to producing a shortlist of likely identities than making a final identification.
%is not yet a high-assurance biometric gate. 
Third, WiMANS cross-environment transfer shows that room-specific multipath remains a major barrier. Fourth, the WiMANS baseline comparison is not a matched-input ablation of every baseline under the same re-extracted complex CSI pipeline. A practical deployment should evaluate across rooms, days, devices, and collection orders, and should combine closed-set identification with calibrated verification or metric-learning losses.

Passive biometric sensing also requires explicit governance. Deployments should require informed enrolment and consent, visible opt-out mechanisms, retention limits for CSI-derived identity data, and local or on-device processing when possible. Passive identification systems such as Argus should not be used as a covert identity gate; the open-set results show that unknown users can still be assigned confident closed-set identities without a separate verification layer.

\section{Conclusion}
We presented Argus, a passive Wi-Fi CSI identification system based on compact statgram inputs and a lightweight decoder-only Transformer. On a 154-subject split, Argus reaches $84.85\% \pm 1.31\%$ Top-1 and $99.26\%$ Top-5 after 60-second aggregation. On WiMANS, it remains within 1.23 points of the strongest baselines on average while using $27\times$ fewer inference FLOPs. Argus shows that compact CSI statistics can support scalable passive identification, but robust deployment still requires stronger open-set rejection, ordering-independent validation, and cross-environment generalization.

\clearpage
\bibliography{references}

\clearpage
\appendix
\section{Hardware Specifications}
The experiments were run on an Apple Mac Studio (Mac16,9) with an Apple M4 Max system-on-chip, 16-core CPU consisting of 12 performance cores and 4 efficiency cores, 40-core integrated Apple GPU, and 128 GB unified memory. Python 3.12.11 and PyTorch 2.7.1 with the Apple Metal Performance Shaders backend enabled (device=mps); CUDA was not used because the experiments were run on Apple Silicon. The notebook records NumPy 2.4.6 and pandas 3.0.3, and uses SciPy, scikit-learn, Matplotlib, Jupyter/Notebook, and IPython kernel support for statistics, metrics, plotting, and notebook execution.

\section{Additional EHealth Details}
EHealth data were collected with a 5 GHz router on channel 36, a laptop client, and a single-antenna Raspberry Pi 4B probe running NEXMON firmware. CSI was extracted across 256 subcarriers, of which 234 were usable. The devices were placed 1 m from the participant, with the router and laptop on opposite sides and the probe equidistant from them. Argus uses AdamW with learning rate $3 \times 10^{-4}$, weight decay $10^{-3}$, cosine annealing, dropout 0.15, label smoothing 0.03, gradient clipping at 1.0, and early stopping with patience 16. A training-history audit found best-validation epochs between 38 and 85 across seeds with flat final validation trends, suggesting the 90-epoch budget is not the main performance limiter.

\begin{table*}[t]
\centering
\small
\caption{EHealth statgram statistic rows. Amplitude and calibrated phase use rows 1--18 and zero-fill the reserved rows. Phase-delta uses rows 1--8 and zero-fills the remaining rows in the reproduction artifact.}
\label{tab:stat-rows}
\begin{tabular}{rll}
\toprule
Row & Name & Definition \\
\midrule
1 & mean & Average value over packets in the window. \\
2 & std & Standard deviation over packets. \\
3 & min & Minimum packet value. \\
4 & max & Maximum packet value. \\
5 & last-first & Last packet value minus first packet value. \\
6 & FFT low & Mean temporal FFT power in the lowest third of non-DC bins. \\
7 & FFT mid & Mean temporal FFT power in the middle third of non-DC bins. \\
8 & FFT high & Mean temporal FFT power in the highest third of non-DC bins. \\
9 & p10 & 10th percentile. \\
10 & p25 & 25th percentile. \\
11 & p50 & Median. \\
12 & p75 & 75th percentile. \\
13 & p90 & 90th percentile. \\
14 & IQR & Interquartile range, $p75-p25$. \\
15 & RMS & Root mean square. \\
16 & ZCR & Fraction of sign changes after median centering. \\
17 & autocorr-1 & Pearson autocorrelation at lag 1 packet. \\
18 & autocorr-window & Pearson autocorrelation at lag $\max(2,\lfloor\text{window rows}/10\rfloor)$. \\
19--20 & reserved & Zero-filled reserved rows. \\
\bottomrule
\end{tabular}
\end{table*}

Table~\ref{tab:stat-rows} enumerates the 20 statistic rows. Frequency rows (FFT low/mid/high) are computed from the temporal real-FFT power of each grouped bin after subtracting the temporal mean; the DC bin is discarded and the remaining bins are split into equal low, middle, and high thirds. All views are robustly normalized per window by subtracting the median and dividing by the interquartile range (floored at $\epsilon = 10^{-4}$), computed over the whole window.

The two datasets differ only in grouping, alignment, and conditioning, not in the statistic battery. On EHealth, the 234 usable subcarriers are averaged into $W = 64$ grouped bins per view, giving the $3 \times 20 \times 64$ statgram. On WiMANS, the 30 Intel 5300 subcarriers are used directly without grouping, and each of the nine transmit-receive links contributes one amplitude and one calibrated-phase channel, giving the $18 \times 20 \times 30$ statgram. Because WiMANS windows contain 3,000 packets versus 50 rows on EHealth, the FFT band-energy rows scale with window length; the WiMANS configuration therefore log-compresses the band energies and standardizes every (channel, statistic row) pair to zero mean and unit variance using training-split statistics only. Without this conditioning, the band-energy rows dominate the un-normalized patch embedding and cost roughly 5 points on the hardest configuration.

Runtime measurements in the artifact use median forward-pass latency after 30 warm-up passes and 120 timed repeats, with accelerator synchronization when available. FLOPs are derived from architecture-level multiply-add counts for patch embedding, attention projections, attention score/value products, feed-forward layers, and the classifier; segment-level compute multiplies the single-window cost by the number of aggregated windows. These deterministic estimates are used for compute comparisons because wall-clock latency can be affected by kernel dispatch and background system load.

\begin{comment}

\begin{figure}[t]
\centering
\includegraphics[width=\columnwidth]{eh_retention}
\caption{Input compression on the 8-patch EHealth statgram.}
\label{fig:eh-ret}
\end{figure}

\begin{figure}[t]
\centering
\includegraphics[width=\columnwidth]{eh_frontier}
\caption{Accuracy-compute frontier on EHealth.}
\label{fig:eh-frontier}
\end{figure}

\begin{figure}[t]
\centering
\includegraphics[width=\columnwidth]{eh_openset}
\caption{Open-set threshold trade-off with 31 withheld identities.}
\label{fig:eh-open}
\end{figure}

\begin{figure}[t]
\centering
\includegraphics[width=\columnwidth]{capacity.png}
\caption{Accuracy versus model size on classroom 2.4 GHz.}
\label{fig:capacity}
\end{figure}

\end{comment}

\begin{figure}[t]
\centering
\includegraphics[width=\columnwidth]{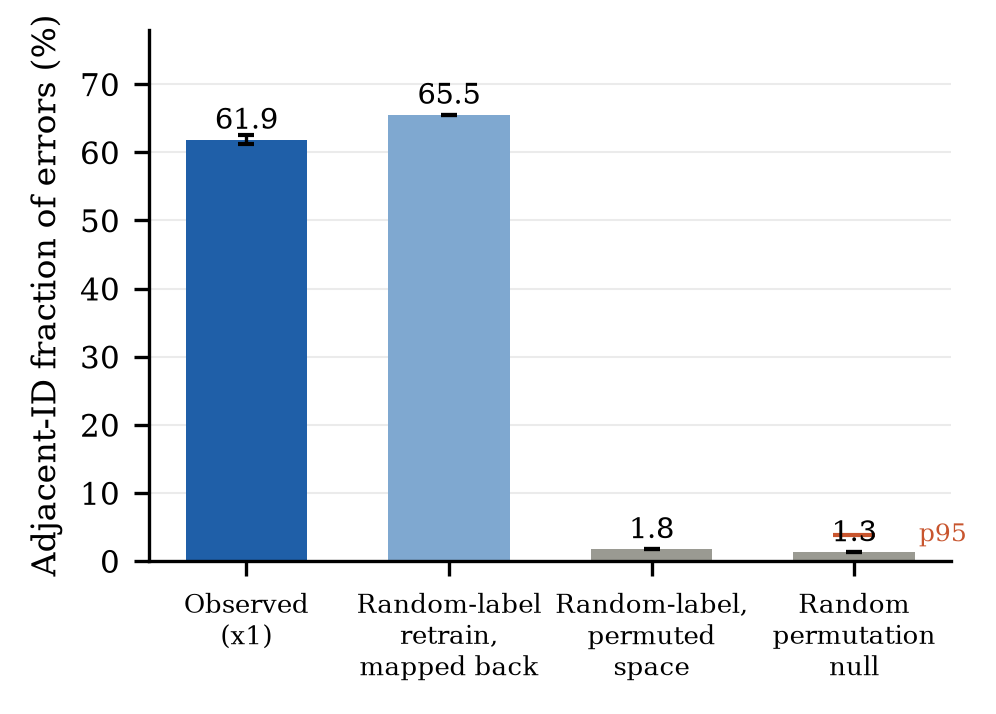}
\caption{Adjacent-ID error structure under relabeling controls.}
\label{fig:eh-adj}
\end{figure}

\section{Additional WiMANS Details}
The final WiMANS configuration uses the same Transformer backbone as EHealth, but with an $18 \times 20 \times 30$ statgram, $4 \times 10$ patches, and a six-way multi-label head trained with weighted binary cross-entropy. The following diagnostics show the effect of the adaptation ladder, token granularity, model capacity, input compression, and cross-environment adaptation.

\begin{figure}[t]
\centering
\includegraphics[width=\columnwidth]{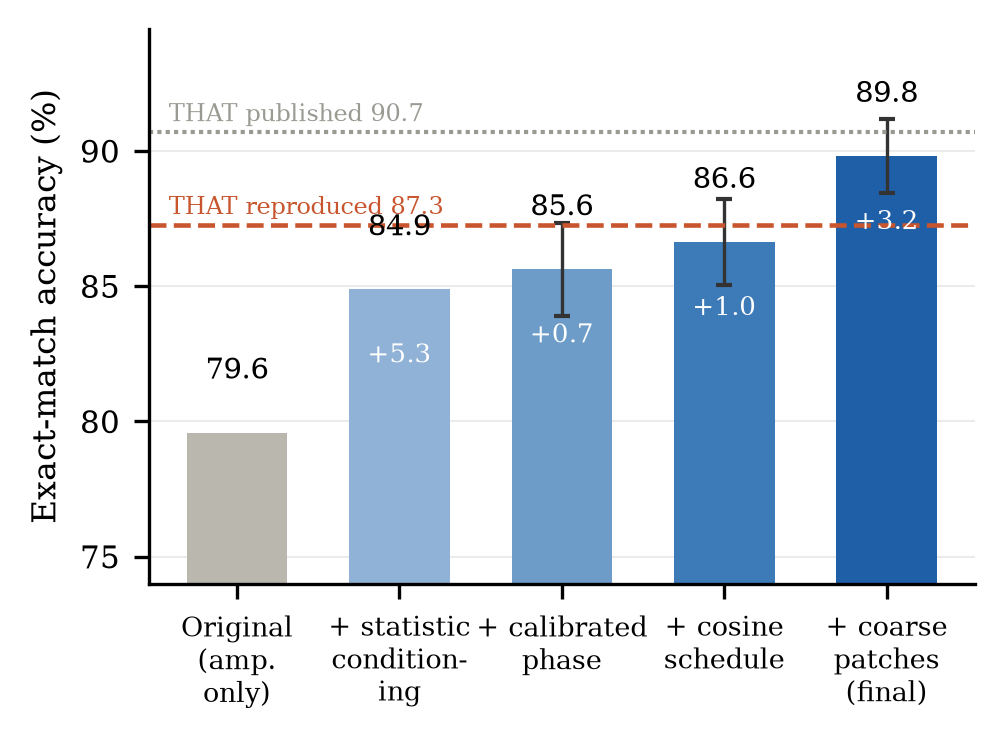}
\caption{Cumulative effect of the WiMANS adaptations on classroom 2.4 GHz.}
\label{fig:ablation_ladder}
\end{figure}

\begin{figure}[t]
\centering
\includegraphics[width=\columnwidth]{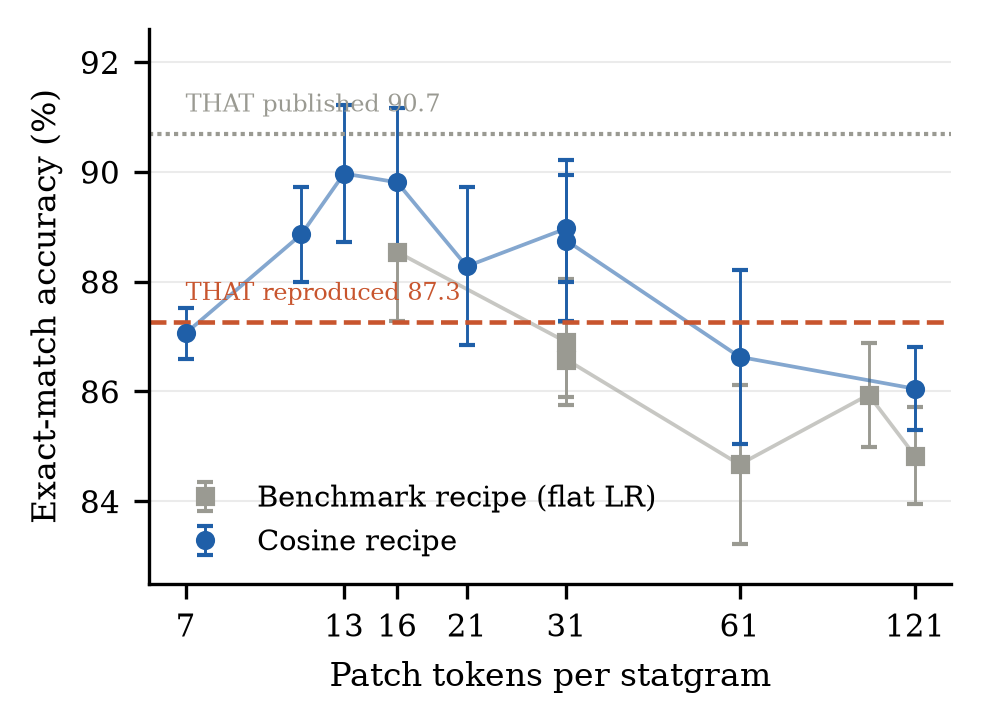}
\caption{Patch-token granularity versus accuracy on classroom 2.4 GHz.}
\label{fig:token_recipe}
\end{figure}

\begin{figure}[t]
\centering
\includegraphics[width=\columnwidth]{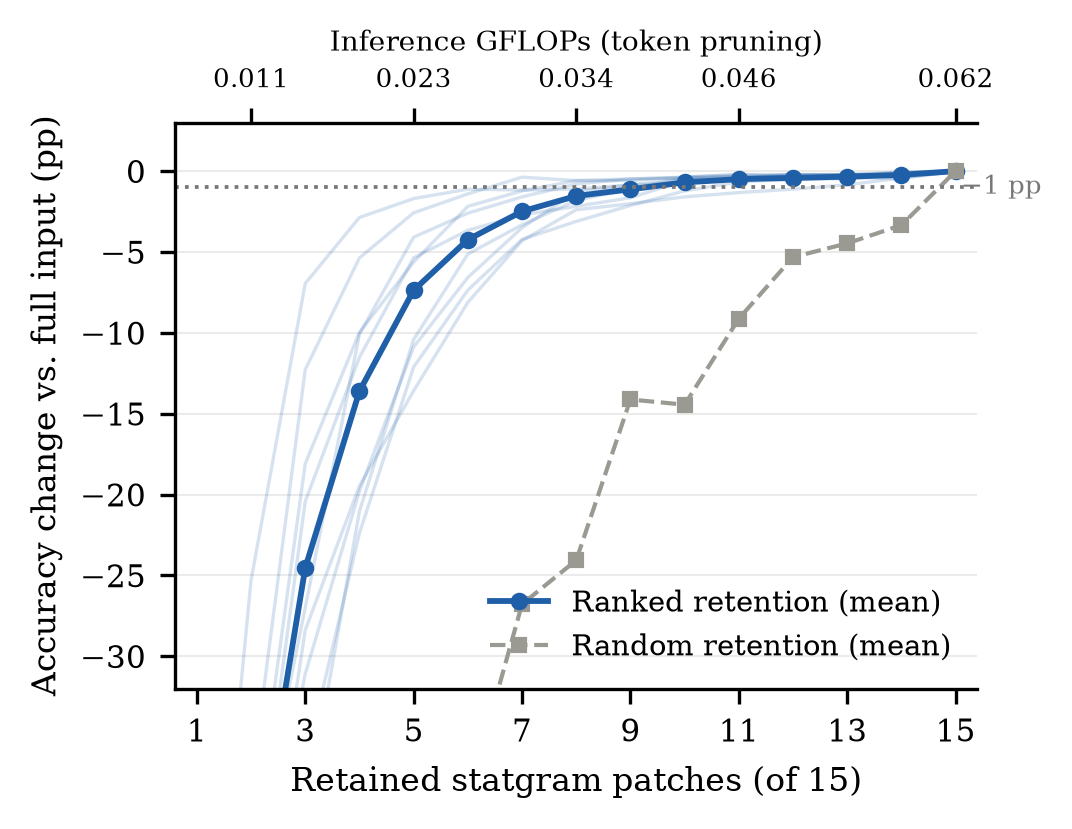}
\caption{Input compression on the final WiMANS model.}
\label{fig:wimans_compression}
\end{figure}

\end{document}